\documentclass[10pt,twocolumn,letterpaper]{article}

\usepackage{times}
\usepackage{epsfig}
\usepackage{subfig}
\usepackage{graphicx}
\usepackage{amsmath}
\usepackage{amssymb}

\usepackage{amsthm}
\usepackage{graphics}
\usepackage{array}
\usepackage{enumerate}

\usepackage[pagebackref=true,breaklinks=true,letterpaper=true,colorlinks,bookmarks=false]{hyperref}
\usepackage{subfloat}

\begin{document}

\title{Deeply Coupled Auto-encoder Networks for\\ Cross-view Classification}

\author{Wen Wang, Zhen Cui, Hong Chang, Shiguang Shan, Xilin Chen\\
Institute of Computing Technology, Chinese Academy of Sciences, Beijing, China\\
{\tt\small \{wen.wang, zhen.cui, hong.chang, shiguang.shan,
xilin.chen\}@vipl.ict.ac.cn} }
\date{November 2013}
\maketitle

\begin{abstract}

The comparison of heterogeneous samples extensively
exists in many applications, especially in the task of image
classification. In this paper, we propose a simple but effective
coupled neural network, called Deeply Coupled Autoencoder
Networks (DCAN), which seeks to build two deep
neural networks, coupled with each other in every corresponding
layers. In DCAN, each deep structure is developed via
stacking multiple discriminative coupled auto-encoders, a
denoising auto-encoder trained with maximum margin criterion
consisting of intra-class compactness and inter-class
penalty. This single layer component makes our model simultaneously
preserve the local consistency and enhance its discriminative
capability. With increasing number of
layers, the coupled networks can gradually narrow the
gap between the two views. Extensive experiments on cross-view
image classification tasks demonstrate the superiority
of our method over state-of-the-art methods.

\end{abstract}

\section{Introduction}

Real-world objects often have different views, which might be endowed with the same semantic. For example, face images can be captured in different poses, which reveal the identity of the same object; images of one face can also be in different modalities, such as pictures under different lighting condition, pose, or even sketches from artists.
In many computer vision applications, such as image retrieval, interests are taken in comparing two types of heterogeneous images, which may come from different views or even different sensors. Since the spanned feature spaces are quite different, it is very difficult to classify these images across views directly. To decrease the discrepancy across views, most of previous works endeavored to learn view-specific linear transforms and to project cross-view samples into a common latent space, and then employed these newly generated features for classification.

Though there are lots of approaches used to learn view-specific projections, they can be divided roughly based on whether the supervised information is used. Unsupervised methods such as Canonical Correlation Analysis (CCA)\cite{hotelling1936relations} and Partial Least Square (PLS) \cite{sharma2011bypassing} are employed to the task of cross-view recognition. Both of them attempt to use two linear mappings to project samples into a common space where the correlation is maximized, while PLS considers the variations rather than only the correlation in the target space. Besides, with use of the mutual information, a Coupled Information-Theoretic Encoding (CITE) method is developed to narrow the inter-view gap for the specific photo-sketch recognition task. And in \cite{wang2012semi}, a semi-coupled dictionary is used to bridge two views. All the methods above consider to reduce the discrepancy between two views, however, the label information is not explicitly taken into account.
With label information available, many methods were further developed to learn a discriminant common space 
For instance, Discriminative Canonical Correlation Analysis (DCCA) \cite{kim2007discriminative} is proposed as an extension of CCA. And In \cite{lin2006inter}, with an additional local smoothness constraints, two linear projections are simultaneously learnt for Common Discriminant Feature Extraction (CDFE). There are also other such methods as the large margin approach \cite{chen2010predictive} and the Coupled Spectral Regression (CSR) \cite{lei2009coupled}.
Recently, multi-view analysis \cite{sharma2012generalized,kan2012multi} is further developed to jointly learn multiple specific-view transforms when multiple views (usually more than 2 views) can be available.

Although the above methods have been extensively applied in the cross-view problem, and have got encouraging performances, they all employed linear transforms to capture the shared features of samples from two views. However, these linear discriminant analysis methods usually depend on the assumption that the data of each class agrees with a Gaussian distribution, while data in real world usually has a much more complex distribution \cite{yan2007graph}. It indicates that linear transforms are insufficient to extract the common features of cross-view images. So it's natural to consider about learning nonlinear features.

A recent topic of interest in nonlinear learning is the research in deep learning. Deep learning attempts to learn nonlinear representations hierarchically via deep structures, and has been applied successfully in many computer vision problems. Classical deep learning methods often stack or compose multiple basic building blocks to yield a deeper structure. See \cite{bengio2013representation} for a recent review of Deep Learning algorithms. Lots of such basic building blocks have been proposed, including sparse coding \cite{lee2007efficient}, restricted Boltzmann machine (RBM) \cite{hinton2006fast}, auto-encoder \cite{hinton2006reducing,bengio2007greedy}, etc. Specifically, the (stacked) auto-encoder has shown its effectiveness in image denoising \cite{xie2012image}, domain adaptation \cite{chen2012marginalized}, audio-visual speech classification \cite{ngiam2011multimodal}, etc.

As we all known, the kernel method, such as Kernel Canonical Correlation Analysis(Kernel CCA) \cite{akaho2006kernel}, is also a widely used approach to learn nonlinear representations. Compared with the kernel method, deep learning is much more flexible and time-saving because the transform is learned rather than fixed and the time needed for training and inference process is beyond the limit of the size of training set.

Inspired by the deep learning works above, we intend to solve the cross-view classification task via deep networks. It's natural to build one single deep neural network with samples from both views, but this kind of network can't handle complex data from totally different modalities and may suffer from inadequate representation capacity. Another way is to learn two different deep neural networks with samples of the different views. However, the two independent networks project samples from different views into different spaces, which makes comparison infeasible. Hence, building two neural networks coupled with each other seems to be a better solution.

In this work, we propose a Deeply Coupled Auto-encoder Networks(DCAN) method that learns the common representations to conduct cross-view classification by building two neural networks deeply coupled respectively, each for one view. We build the DCAN by stacking multiple discriminative coupled auto-encoders, a denoising auto-encoder with maximum margin criterion. The discriminative coupled auto-encoder has a similar input corrupted and reconstructive error minimized mechanism with the denoising auto-encoder proposed in \cite{vincent2008extracting}, but is modified by adding a maximum margin criterion. This kind of criterion has been used in previous works, like \cite{li2006efficient,wang2007feature,zhao2008maximum}, etc. Note that the counterparts from two views are added into the maximum margin criterion simultaneously since they both come from the same class, which naturally couples the corresponding layer in two deep networks. A schematic illustration can be seen in Fig.\ref{fig:frame1}.

The proposed DCAN is related to Multimodal Auto-encoders \cite{ngiam2011multimodal}, Multimodal Restricted Boltzmann Machines and Deep Canonical Correlation Analysis \cite{andrewdeep}. The first two methods tend to learn a single network with one or more layers connected to both views and to predict one view from the other view, and the Deep Canonical Correlation Analysis build two deep networks, each for one view, and only representations of the highest layer are constrained to be correlated. Therefore, the key difference is that we learn two deep networks coupled with each other in representations in each layer, which is of great benefits because the DCAN not only learn two separate deep encodings but also makes better use of data from the both two views. What's more, these differences allow for our model to handle the recognition task even when data is impure and insufficient.

The rest of this paper is organized as follows. Section \ref{sec:DCAN} details the formulation and solution to the proposed Deeply Coupled Auto-encoder Networks. Experimental results in Section \ref{sec:exp} demonstrate the efficacy of the DCAN. In section \ref{sec:Conclusion} a conclusion is given.


\section{Deeply Coupled Auto-encoder Networks}\label{sec:DCAN}

In this section, we first present the basic idea. The second part gives a detailed description of the discriminative coupled auto-encoder. Then, we describe how to stack multiple layers to build a deep network. Finally, we briefly describe the optimization of the model.

\subsection{Basic Idea}

\begin{figure*}
\begin{center}
    \includegraphics[width=0.8\linewidth]{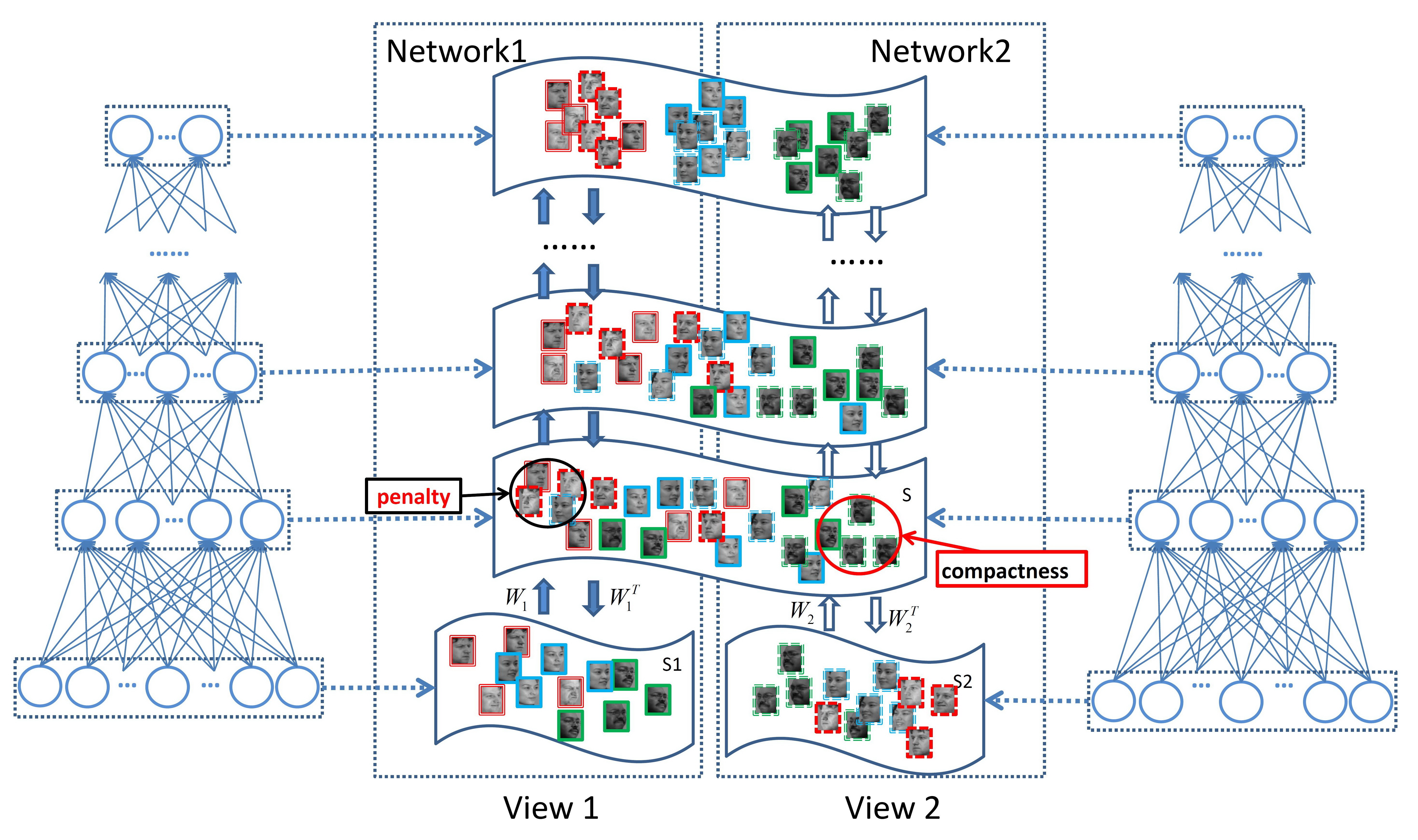}
\end{center}
   \caption{An illustration of our proposed DCAN. The left-most and right-most schematic show the structure of the two coupled network respectively. And the schematic in the middle illustrates how the whole network gradually enhances the separability with increasing layers, where pictures with solid line border denote samples from view 1, those with dotted  line border denote samples from view 2, and different colors imply different subjects.
   }
\label{fig:frame1}
\end{figure*}

As shown in Fig.\ref{fig:frame1}, the Deeply Coupled Auto-encoder Networks(DCAN) consists of two deep networks coupled with each other, and each one is for one view. The network structures of the two deep networks are just like the left-most and the right-most parts in Fig.\ref{fig:frame1}, where circles means the units in each layers (pixels in a input image for the input layer and hidden representation in higher layers), and arrows denote the full connections between adjacent layers. And the middle part of Fig.\ref{fig:frame1} illustrates how the whole network projects samples in different views into a common space and gradually enhances the separability with increasing layers.

The two deep networks are both built through stacking multiple similar coupled single layer blocks because a single coupled layer might be insufficient, and the method of stacking multiple layers and training each layer greedily has be proved efficient in lots of previous works, such as those in \cite{hinton2006reducing,bengio2007greedy}. With the number of layers increased, the whole network can compactly represent a significantly larger set of transforms than shallow networks , and gradually narrow the gap with the discriminative capacity enhanced.

We use a discriminative coupled auto-encoders trained with maximum margin criterion as a single layer component. Concretely, we incorporate the additional noises in the training process while maximizing the margin criterion, which makes the learnt mapping more stable as well as discriminant. Note that the maximum margin criterion also works in coupling two corresponding layers. Formally, the discriminative coupled auto-encoder can be written as follows:

\begin{eqnarray}\label{equ: object_1}
&& \quad\min_{f_x,f_y}\quad L(X,f_x)+L(Y,f_y)\label{equ: object_1}\\
&&s.t. \quad G_1(H_x,H_y) - G_2(H_x,H_y)\leq \varepsilon, \label{equ_obj1_st}
\end{eqnarray}
where $X,Y$ denote inputs from the two views, and $H_x,H_y$ denote hidden representations of the two views respectively. $f_x:X\longrightarrow H_x,f_y:Y\longrightarrow H_y $ are the transforms we intend to learn,  and we denote the reconstructive error as $L(\cdot)$, and maximum margin criterion as $G_1(\cdot)-G_2(\cdot)$, which are described detailedly in the next subsection.$\varepsilon$ is the threshold of the maximum margin criterion.

\subsection{Discriminative coupled auto-encoder}\label{subsec:layer}

In the problem of cross-view, there are two types of heterogenous samples. Without loss of generality, we denote samples from one view as $X=[x_1,\cdots,x_{n}]$ , and those from the other view as $Y=[y_1,\cdots,y_{n}]$, in which $n$ is the sample sizes. Noted that the corresponding labels are known, and $H_x,H_y$ denote hidden representations of the two views we want to learn.

The DCAN attempts to learn two nonlinear transforms $f_x:X \longrightarrow H_x$ and $f_y:Y \longrightarrow H_y$ that can project the samples from two views to one discriminant common space respectively, in which the local neighborhood relationship as well as class separability should be well preserved for each view.
The auto-encoder like structure stands out in preserving the local consistency, and the denoising form enhances the robustness of learnt representations. However, the discrimination isn't taken into consideration. Therefore, we modify the denoising auto-encoder by adding a maximum margin criterion consisting of intra-class compactness and inter-class penalty. And the best nonlinear transformation is a trade-off between local consistency preserving and separability enhancing.

Just like the one in denoising auto-encoder, the reconstructive error $L(\cdot)$ in Eq.(\ref{equ: object_1}) is formulated as follows:

\begin{eqnarray}
L(X,\Theta) = \sum_{x\in{X^p}}{\mathbb{E}_{\tilde{x}\sim{P(\tilde{x}|x)}}} \|\hat{x}-x\|\\
L(Y,\Theta) = \sum_{y\in{Y^p}}{\mathbb{E}_{\tilde{y}\sim{P(\tilde{y}|y)}}} \|\hat{y}-y\|
\end{eqnarray}

where $\mathbb{E}$ calculates the expectation over corrupted versions $\tilde{X},\tilde{Y}$ of
examples $X,Y$ obtained from a corruption process $P(\tilde{x}|x),P(\tilde{y}|y)$. $\Theta=\{W_x,W_y,b_x,b_y,c_x,c_y\}$ specifies the two nonlinear transforms $f_x,f_y$ , where $W_x,W_y$ is the weight matrix, and $b_x,b_y,c_x,c_y$ are the bias of encoder and decoder respectively, and $\hat{X},\hat{Y}$ are calculated through the decoder process :

\begin{equation}\label{equ:decoder}
\begin{split}
  \hat{X} = s(W_x^T H_x + c_x)\\
  \hat{Y} = s(W_y^T H_y + c_y)
\end{split}
\end{equation}

And hidden representations $H_x,H_y$ are obtained from the encoder that is a similar mapping with the decoder,

\begin{equation}\label{equ:encoder}
\begin{split}
  H_x = s(W_x \tilde{X} + b_x)\\
  H_y = s(W_y \tilde{Y} + b_y)
\end{split}
\end{equation}

where $s$ is the nonlinear activation function, such as the point-wise hyperbolic tangent operation on linear projected features, \textit{i.e.},
\begin{equation}
  s(x)=\frac{e^{ax}-e^{-ax}}{e^{ax}+e^{-ax}}
\end{equation}
in which $a$ is the gain parameter.

Moreover, for the maximum margin criterion consisting of intra-class compactness and inter-class penalty, the constraint term $G_1(\cdot)-G_2(\cdot)$ in Eq.(\ref{equ: object_1}) is used to realize coupling since samples of the same class are treated similarly no matter which view they are from.

Assuming $S$ is the set of sample pairs from the same class, and $D$ is the set of sample pairs from different classes.  Note that the counterparts from two views are naturally added into $S,D$ since it's the class rather than the view that are considered.

Then, we characterize the compactness as follows,
\begin{eqnarray}
& G_1(H) = \frac{1}{2N_1}\sum\limits_{I_i,I_j\in{S}}\|h_i-h_j\|^2, \label{equ:G1}
\end{eqnarray}
where $h_i$ denotes the corresponding hidden representation of an input $I_i\in{X\bigcap{Y}}$ and is a sample from either view 1 or view 2, and $N_1$ is the size of $S$.

Meanwhile, the goal of the inter-class separability is to push the adjacent samples from different classes far away, which can be formulated as follows,
\begin{eqnarray}
& G_2(H) =\frac{1}{2N_2}\sum\limits_{\tiny\substack{I_i,I_j\in{D}\\I_j\in{KNN(I_i)}}}\|h_i-h_j\|^2, \label{equ:G2}
\end{eqnarray}
where $I_j$ belongs to the $k$ nearest neighbors of $I_i$ with different class labels, and $N_2$ is the number of all pairs satisfying the condition.

And the function of $G_1(H),G_2(H)$ is illustrated in the middel part of Fig.\ref{fig:frame1}. In the projected common space denoted by $S$, the compactness term $G_1(\cdot)$ shown by red ellipse works by pulling intra-class samples together while the penalty term $G_2(\cdot)$ shown by black ellipse tend to push adjacent inter-class samples away.

Finally, by solving the optimization problem Eq.(\ref{equ: object_1}), we can learn a couple of nonlinear transforms $f_x,f_y$ to transform the original samples from both views into a common space.

\subsection{Stacking coupled auto-encoder}

Through the training process above, we model the map between original sample space and a preliminary discriminant subspace with gap eliminated, and build a hidden representation $H$ which is a trade-off between approximate preservation on local consistency and the distinction of the projected data. But since real-world data is highly complicated, using a single coupled layer to model the vast and complex real scenes might be insufficient. So we choose to stack multiple such coupled network layers described in subsection \ref{subsec:layer}. With the number of layers increased, the whole network can compactly represent a significantly larger set of transforms than shallow networks, and gradually narrow the gap with the discriminative ability enhanced.

Training a deep network with coupled nonlinear transforms can be achieved by the canonical greedy layer-wise approach \cite{hinton2006fast,bengio2007greedy}. Or to be more precise, after training a single layer coupled network, one can compute a new feature $H$ by the encoder in Eq.(\ref{equ:encoder}) and then feed it into the next layer network as the input feature. In practice, we find that stacking multiple such layers can gradually reduce the gap and improve the recognition performance (see Fig.\ref{fig:frame1} and Section \ref{sec:exp}).

\subsection{Optimization}

We adopt the Lagrangian multiplier method to solve the objective function Eq.(\ref{equ: object_1}) with the constraints Eq.(\ref{equ_obj1_st}) as follows:

\begin{equation}\label{equ: object_3}
    \begin{split}
        \min_{\Theta} \quad &\lambda(L(X,\Theta)+L(Y,\Theta)) + (G_1(H)-G_2(H)) + \\
        &\gamma(\frac{1}{2}\|W_x\|_F^2+\frac{1}{2}\|W_y\|_F^2)
    \end{split}
\end{equation}

where the first term is the the reconstruction error, the second term is the maximum margin criterion, and the last term is the shrinkage constraints called the Tikhonov regularizers in \cite{trevor2001elements}, which is utilized to decrease the magnitude of the weights and further to help prevent over-fitting. $\lambda$ is the balance parameter between the local consistency and empirical separability. And $\gamma$ is called the weight decay parameter and is usually set to a small value, \textit{e.g.}, 1.0e-4.

To optimize the objective function (\ref{equ: object_3}), we use back-propagation to calculate the gradient and then employ the limited-memory BFGS (L-BFGS) method \cite{nocedal2006numerical,le2011optimization}, which is often used to solve nonlinear optimization problems without any constraints. L-BFGS is particularly suitable for problems with a large amount of variables under the moderate memory requirement. To utilize L-BFGS, we need to calculate the gradients of the object function. Obviously, the object function in (\ref{equ: object_3}) is differential to these parameters $\Theta$, and we use Back-propagation \cite{lecun1998efficient} method to derive the derivative of the overall cost function. In our setting, we find the objective function can achieve as fast convergence as described in \cite{le2011optimization}.

\section{Experiments}\label{sec:exp}

In this section, the proposed DCAN is evaluated on two datasets, Multi-PIE \cite{gross2007cmu} and CUHK Face Sketch FERET (CUFSF) \cite{zhang2011coupled,wang2009face}.

\subsection{Databases}

\textbf{Multi-PIE} dataset \cite{gross2007cmu} is employed to evaluate face recognition across pose. Here a subset from the 337 subjects in 7 poses ($-45^\circ, -30^\circ, -15^\circ, 0^\circ, 15^\circ, 30^\circ, 45^\circ$), 3 expression (Neutral,Smile, Disgust), no flush illumination from 4 sessions are selected to validate our method. We randomly choose 4 images for each pose of each subject, then randomly partition the data into two parts: the training set with 231 subjects (\textit{i.e.}, $231\times 7\times 4=6468$ images) and the testing set with the rest subjects.

\textbf{CUHK Face Sketch FERET (CUFSF)} dataset \cite{zhang2011coupled,wang2009face} contains two types of face images: photo and sketch. Total 1,194 images (one image per subject) were collected with lighting variations from FERET dataset \cite{phillips1998feret}. For each subject, a sketch is drawn with shape exaggeration. According to the configuration of \cite{kan2012multi}, we use the first 700 subjects as the training data and the rest subjects as the testing data.

\subsection{Settings}

All images from Multi-PIE and CUFSF are cropped into 64$\times$80 pixels without any preprocess. We compare the proposed DCAN method with several baselines and state-of-the-art methods, including CCA \cite{hotelling1936relations}, Kernel CCA \cite{akaho2006kernel}, Deep CCA \cite{andrewdeep}, FDA \cite{belhumeur1997eigenfaces}, CDFE \cite{lin2006inter}, CSR \cite{lei2009coupled}, PLS \cite{sharma2011bypassing} and MvDA \cite{kan2012multi}. The first seven methods are pairwise methods for cross-view classification. MvDA jointly learns all transforms when multiple views can be utilized, and has achieved the state-of-the-art results in their reports \cite{kan2012multi}.

The Principal Component Analysis (PCA) \cite{belhumeur1997eigenfaces} is used for dimension reduction. In our experiments, we set the default dimensionality as 100 with preservation of most energy except Deep CCA, PLS, CSR and CDFE, where the dimensionality are tuned in [50,1000] for the best performance. For all these methods, we report the best performance by tuning the related parameters according to their papers. Firstly, for Kernel CCA, we experiment with Gaussian kernel and polynomial kernel and adjust the parameters to get the best performance. Then for Deep CCA \cite{andrewdeep}, we strictly follow their algorithms and tune all possible parameters, but the performance is inferior to CCA. One possible reason is that Deep CCA only considers the correlations on training data (as reported in their paper) so that the learnt mode overly fits the training data, which thus leads to the poor generality on the testing set. Besides, the parameter $\alpha$ and $\beta$ are respectively traversed in [0.2,2] and [0.0001,1] for CDFE, the parameter $\lambda$ and $\eta$ are searched in [0.001,1] for CSR, and the reduced dimensionality is tuned for CCA, PLS, FDA and MvDA.

\begin{figure}
\begin{center}
\subfloat[]{\label{fig:lambda}
\includegraphics[height=3cm]{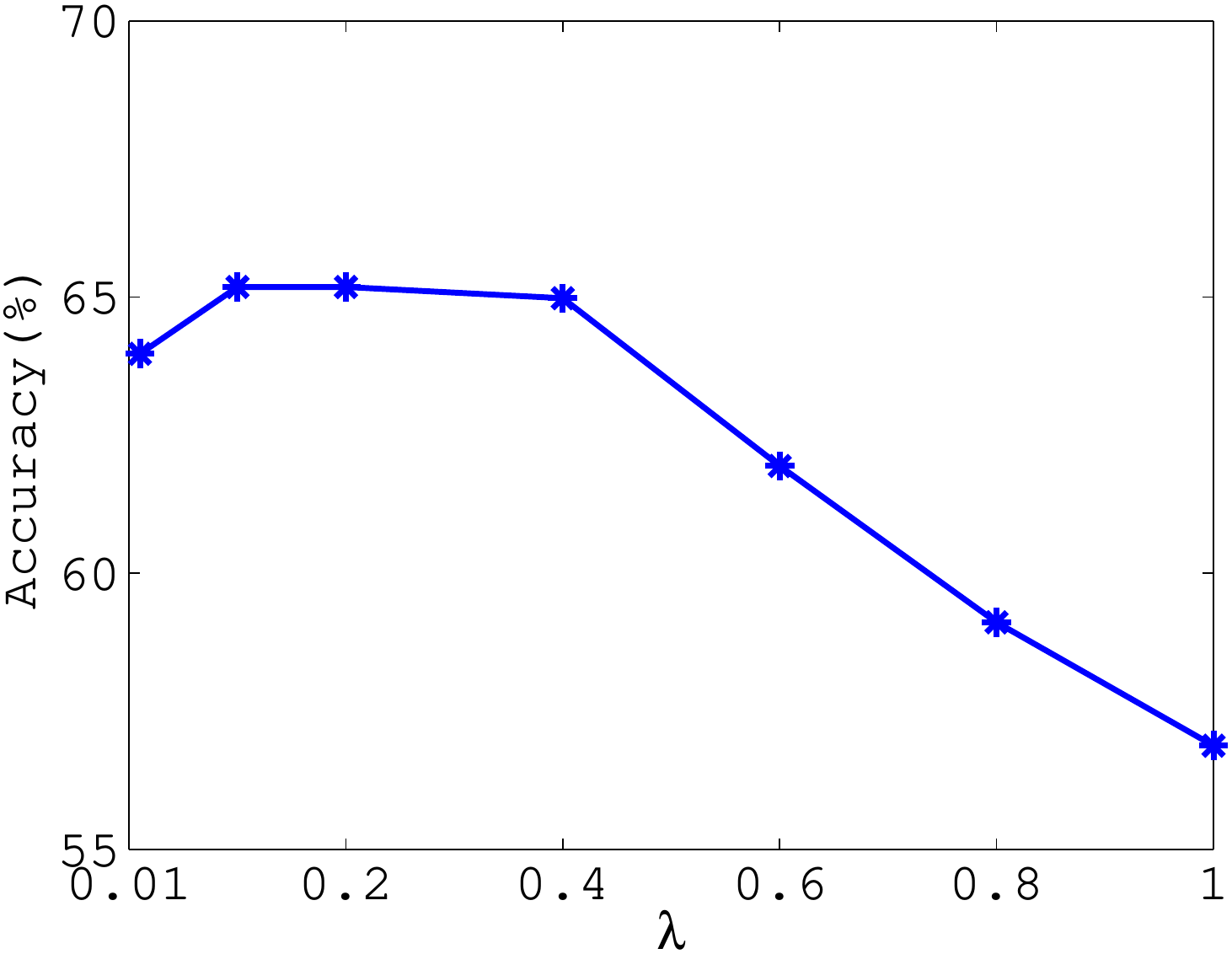}
}
\subfloat[]{\label{fig:knn}
\includegraphics[height=3cm]{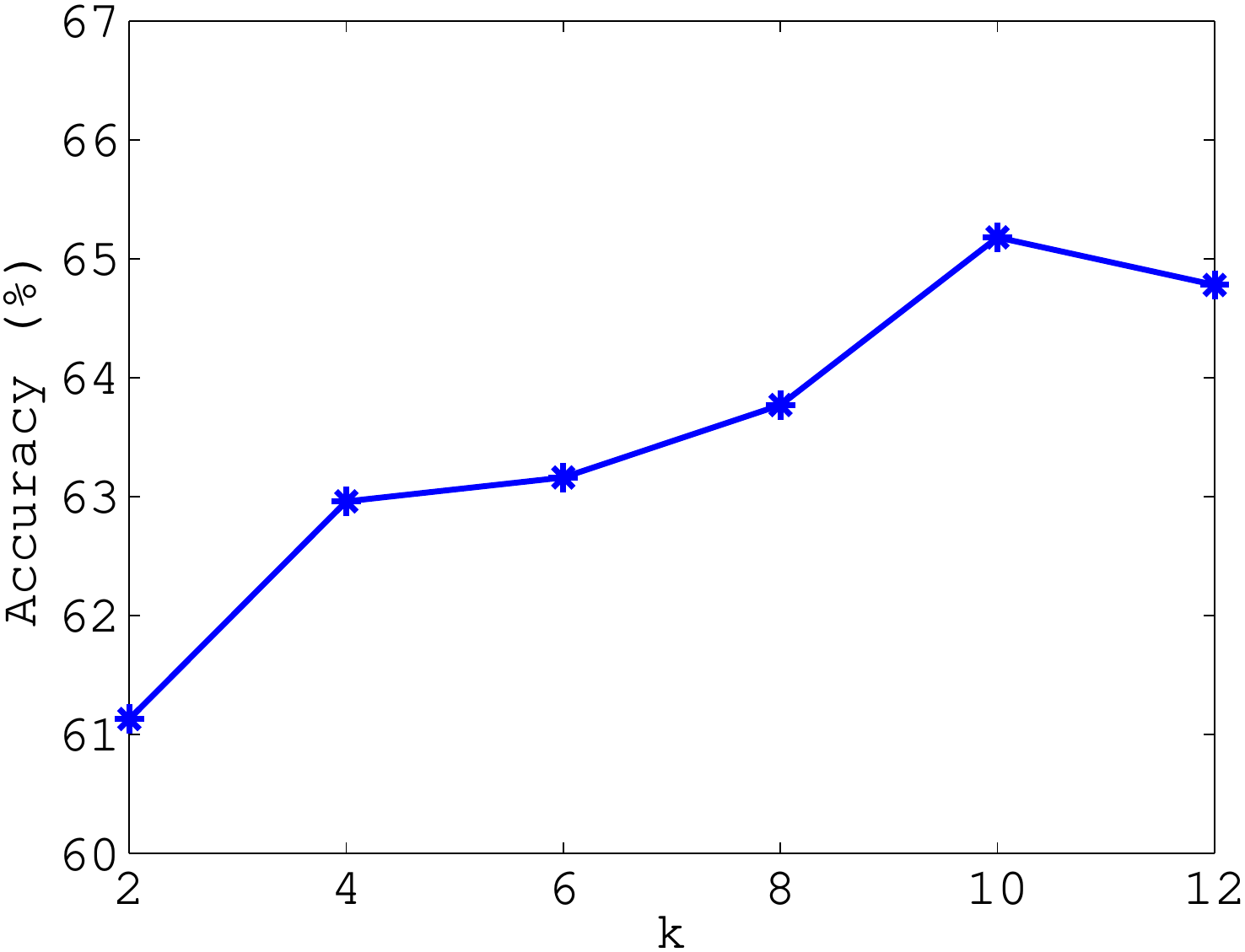}
}
\end{center}

\caption{ The performance with varied parameter values for our proposed DCAN. The sketch and photo images in CUFSF \cite{zhang2011coupled,wang2009face} are respectively used for the gallery and probe set. (a) Varied $\lambda$ with fixed $k=10$. (b) Varied $k$ with fixed $\lambda=0.2$.
} \label{fig:parameters}
\end{figure}

As for our proposed DCAN, the performance on CUFSF database of varied parameters, $\lambda,k$, is shown in Fig.\ref{fig:parameters}. In following experiments, we set $\lambda=0.2, \gamma=1.0e-4$, $k=10$ and $a=1$. With increasing layers, the number of hidden neurons are gradually reduced by $10$, \emph{i.e.,} $90,80,70,60$ if four layers.

\begin{table}\footnotesize
\setlength{\tabcolsep}{8pt}
\begin{center}
{
\begin{tabular}{|p{60pt}<{\centering}|p{60pt}<{\centering}|}
    \hline
    Method & Accuracy\\
    \hline\hline\centering
    CCA\cite{hotelling1936relations} & 0.698 \\
    \hline
    KernelCCA\cite{hardoon2004canonical} & 0.840\\
    \hline
    DeepCCA\cite{andrewdeep} & 0.599\\
    \hline
    FDA\cite{belhumeur1997eigenfaces} & 0.814\\
    \hline
    CDFE\cite{lin2006inter} & 0.773 \\
    \hline
    CSR\cite{lei2009coupled} & 0.580\\
    \hline
    PLS\cite{sharma2011bypassing} & 0.574\\
    \hline
    MvDA\cite{kan2012multi} & 0.867\\
    \hline\hline
    \textbf{DCAN-1} & 0.830\\
    \hline
    \textbf{DCAN-2} & 0.877\\
    \hline
    \textbf{DCAN-3} & \textbf{0.884}\\
    \hline
    \textbf{DCAN-4} & 0.879    \\
    \hline
\end{tabular}
}
\end{center}
\caption{Evaluation on Multi-PIE database in terms of mean accuracy. DCAN-k means a stacked k-layer network.}
\label{tab:Mean_MPIE}
\end{table}

\begin{table*}
\footnotesize
\setlength{\tabcolsep}{1.5pt}
\centering
    \subfloat[CCA, $Ave=0.698$]
    {
    \begin{tabular}{|c|ccccccc|}
    \hline
      &$-45^\circ$ &$-30^\circ$ &$-15^\circ$ &$0^\circ$ &$15^\circ$ &$30^\circ$ &$45^\circ$ \\
    \hline\hline
    $-45^\circ$ &1.000 & 0.816 & 0.588 & 0.473 & 0.473 & 0.515 & 0.511 \\
    $-30^\circ$ & 0.816 &1.000 & 0.858 & 0.611 & 0.664 & 0.553 & 0.553 \\
    $-15^\circ$ & 0.588 & 0.858 &1.000 & 0.894 & 0.807 & 0.602 & 0.447 \\
    $0^\circ$ & 0.473 & 0.611 & 0.894 &1.000 & 0.909 & 0.604 & 0.484 \\
    $15^\circ$ & 0.473 & 0.664 & 0.807 & 0.909 &1.000 & 0.874 & 0.602 \\
    $30^\circ$ & 0.515 & 0.553 & 0.602 & 0.604 & 0.874 &1.000 & 0.768 \\
    $45^\circ$ & 0.511 & 0.553 & 0.447 & 0.484 & 0.602 & 0.768 &1.000 \\
    \hline
    \end{tabular}
    \label{tab:CCA}
    }
\centering
    \subfloat[KernelCCA, $Ave=0.840$]
    {
    \begin{tabular}{|c|rrrrrrr|}
    \hline
      &$-45^\circ$ &$-30^\circ$ &$-15^\circ$ &$0^\circ$ &$15^\circ$ &$30^\circ$ &$45^\circ$ \\
    \hline\hline
    $-45^\circ$ & 1.000 & 0.878 & 0.810 & 0.756 & 0.706 & 0.726 & 0.737 \\
    $-30^\circ$ & 0.878 & 1.000 & 0.892 & 0.858 & 0.808 & 0.801 & 0.757 \\
    $-15^\circ$ & 0.810 & 0.892 & 1.000 & 0.911 & 0.880 & 0.861 & 0.765 \\
    $0^\circ$   & 0.756 & 0.858 & 0.911 & 1.000 & 0.938 & 0.759 & 0.759 \\
    $15^\circ$  & 0.706 & 0.808 & 0.880 & 0.938 & 1.000 & 0.922 & 0.845 \\
    $30^\circ$  & 0.726 & 0.801 & 0.861 & 0.759 & 0.922 & 1.000 & 0.912 \\
    $45^\circ$  & 0.737 & 0.757 & 0.765 & 0.759 & 0.845 & 0.912 & 1.000 \\
    \hline
    \end{tabular}
    \label{tab:KernelCCA}
    }
\quad
\centering
    \subfloat[DeepCCA, $Ave=0.599$]
    {
    \begin{tabular}{|c|rrrrrrr|}
    \hline
      &$-45^\circ$ &$-30^\circ$ &$-15^\circ$ &$0^\circ$ &$15^\circ$ &$30^\circ$ &$45^\circ$ \\
    \hline\hline
    $-45^\circ$ & 1.000 & 0.854 & 0.598 & 0.425 & 0.473 & 0.522 & 0.523 \\
    $-30^\circ$ & 0.854 & 1.000 & 0.844 & 0.578 & 0.676 & 0.576 & 0.566 \\
    $-15^\circ$ & 0.598 & 0.844 & 1.000 & 0.806 & 0.807 & 0.602 & 0.424 \\
    $0^\circ$   & 0.425 & 0.578 & 0.806 & 1.000 & 0.911 & 0.599 & 0.444 \\
    $15^\circ$  & 0.473 & 0.676 & 0.807 & 0.911 & 1.000 & 0.866 & 0.624 \\
    $30^\circ$  & 0.522 & 0.576 & 0.602 & 0.599 & 0.866 & 1.000 & 0.756 \\
    $45^\circ$  & 0.523 & 0.566 & 0.424 & 0.444 & 0.624 & 0.756 & 1.000 \\
    \hline
    \end{tabular}
    \label{tab:DeepCCA}
    }
\centering
    \subfloat[FDA, $Ave=0.814$]
    {
    \begin{tabular}{|c|rrrrrrr|}
    \hline
      &$-45^\circ$ &$-30^\circ$ &$-15^\circ$ &$0^\circ$ &$15^\circ$ &$30^\circ$ &$45^\circ$ \\
    \hline\hline
    $-45^\circ$ &1.000 & 0.847 & 0.754 & 0.686 & 0.573 & 0.610 & 0.664 \\
    $-30^\circ$ & 0.847 &1.000 & 0.911 & 0.847 & 0.807 & 0.766 & 0.635 \\
    $-15^\circ$ & 0.754 & 0.911 &1.000 & 0.925 & 0.896 & 0.821 & 0.602 \\
    $0^\circ$ & 0.686 & 0.847 & 0.925 &1.000 & 0.964 & 0.872 & 0.684 \\
    $15^\circ$ & 0.573 & 0.807 & 0.896 & 0.964 &1.000 & 0.929 & 0.768 \\
    $30^\circ$ & 0.610 & 0.766 & 0.821 & 0.872 & 0.929 &1.000 & 0.878 \\
    $45^\circ$ & 0.664 & 0.635 & 0.602 & 0.684 & 0.768 & 0.878 &1.000 \\
    \hline
    \end{tabular}
    \label{tab:FDA}
    }
\quad
\centering
    \subfloat[CDFE, $Ave=0.773$]
    {
    \begin{tabular}{|c|rrrrrrr|}
    \hline
    &$-45^\circ$ &$-30^\circ$ &$-15^\circ$ &$0^\circ$ &$15^\circ$ &$30^\circ$ &$45^\circ$ \\
    \hline\hline
    $-45^\circ$ &1.000 & 0.854 & 0.714 & 0.595 & 0.557 & 0.633 & 0.608 \\
    $-30^\circ$ & 0.854 &1.000 & 0.867 & 0.746 & 0.688 & 0.697 & 0.606 \\
    $-15^\circ$ & 0.714 & 0.867 &1.000 & 0.887 & 0.808 & 0.704 & 0.579 \\
    $0^\circ$ & 0.595 & 0.746 & 0.887 &1.000 & 0.916 & 0.819 & 0.651 \\
    $15^\circ$ & 0.557 & 0.688 & 0.808 & 0.916 &1.000 & 0.912 & 0.754 \\
    $30^\circ$ & 0.633 & 0.697 & 0.704 & 0.819 & 0.912 &1.000 & 0.850 \\
    $45^\circ$ & 0.608 & 0.606 & 0.579 & 0.651 & 0.754 & 0.850 &1.000 \\
    \hline
    \end{tabular}\label{tab:CDFE}
    }
\centering
    \subfloat[MvDA, $Ave=0.867$]
    {
    \begin{tabular}{|c|rrrrrrr|}
    \hline
      &$-45^\circ$ &$-30^\circ$ &$-15^\circ$ &$0^\circ$ &$15^\circ$ &$30^\circ$ &$45^\circ$ \\
    \hline\hline
    $-45^\circ$ &1.000 & 0.914 & 0.854 & 0.763 & 0.710 & 0.770 & 0.759 \\
    $-30^\circ$ & 0.914 &1.000 & 0.947 & 0.858 & 0.812 & 0.861 & 0.766 \\
    $-15^\circ$ & 0.854 & 0.947 &1.000 & 0.923 & 0.880 & 0.894 & 0.775 \\
    $0^\circ$ & 0.763 & 0.858 & 0.923 &1.000 & 0.938 & 0.900 & 0.750 \\
    $15^\circ$ & 0.710 & 0.812 & 0.880 & 0.938 &1.000 & 0.923 & 0.807 \\
    $30^\circ$ & 0.770 & 0.861 & 0.894 & 0.900 & 0.923 &1.000 & 0.934 \\
    $45^\circ$ & 0.759 & 0.766 & 0.775 & 0.750 & 0.807 & 0.934 &1.000 \\
    \hline
    \end{tabular}\label{tab:MvDA}
    }
\quad
\centering
    \subfloat[DCAN-1, $Ave=0.830$]
    {
    \begin{tabular}{|c|rrrrrrr|}
    \hline
    &$-45^\circ$ &$-30^\circ$ &$-15^\circ$ &$0^\circ$ &$15^\circ$ &$30^\circ$ &$45^\circ$ \\
    \hline\hline
    $-45^\circ$ &1.000 & 0.872 & 0.819 & 0.730 & 0.655 & 0.708 & 0.686 \\
    $-30^\circ$ & 0.856 &1.000 & 0.881 & 0.825 & 0.754 & 0.737 & 0.650 \\
    $-15^\circ$ & 0.807 & 0.874 &1.000 & 0.869 & 0.865 & 0.781 & 0.681 \\
    $0^\circ$ & 0.757 & 0.854 & 0.896 &1.000 & 0.938 & 0.858 & 0.790 \\
    $15^\circ$ & 0.688 & 0.777 & 0.854 & 0.916 &1.000 & 0.900 & 0.823 \\
    $30^\circ$ & 0.708 & 0.735 & 0.788 & 0.834 & 0.918 &1.000 & 0.916 \\
    $45^\circ$ & 0.719 & 0.715 & 0.697 & 0.752 & 0.832 & 0.909 &1.000 \\
    \hline
    \end{tabular}\label{tab:DCAN_1}
    }
\centering
    \subfloat[DCAN-3, $Ave=0.884$]
    {
    \begin{tabular}{|c|rrrrrrr|}
    \hline
    &$-45^\circ$ &$-30^\circ$ &$-15^\circ$ &$0^\circ$ &$15^\circ$ &$30^\circ$ &$45^\circ$ \\
    \hline\hline
    $-45^\circ$ &1.000 & 0.905 & 0.876 & 0.783 & 0.714 & 0.779 & 0.796 \\
    $-30^\circ$ & 0.927 &1.000 & 0.954 & 0.896 & 0.850 & 0.825 & 0.730 \\
    $-15^\circ$ & 0.867 & 0.929 &1.000 & 0.905 & 0.905 & 0.867 & 0.757 \\
    $0^\circ$ & 0.832 & 0.876 & 0.925 &1.000 & 0.958 & 0.896 & 0.808 \\
    $15^\circ$ & 0.765 & 0.865 & 0.907 & 0.951 &1.000 & 0.929 & 0.874 \\
    $30^\circ$ & 0.779 & 0.832 & 0.870 & 0.916 & 0.945 &1.000 & 0.949 \\
    $45^\circ$ & 0.794 & 0.777 & 0.785 & 0.812 & 0.876 & 0.938 &1.000 \\
    \hline
    \end{tabular}\label{tab:DCAN_3}
    }
\caption{Results of CCA, FDA \cite{belhumeur1997eigenfaces}, CDFE \cite{lin2006inter}, MvDA \cite{kan2012multi} and DCAN on MultiPIE dataset in terms of rank-1 recognition rate. DCAN-k means a stacked k-layer network. Due to space limitation, the results of other methods cannot be reported here, but their mean accuracies are shown in Table \ref{tab:Mean_MPIE}.}
\label{tab:DCAN}
\end{table*}

\subsection{Face Recognition across Pose}

\begin{figure}[t]
\begin{center}
    \includegraphics[height=6cm]{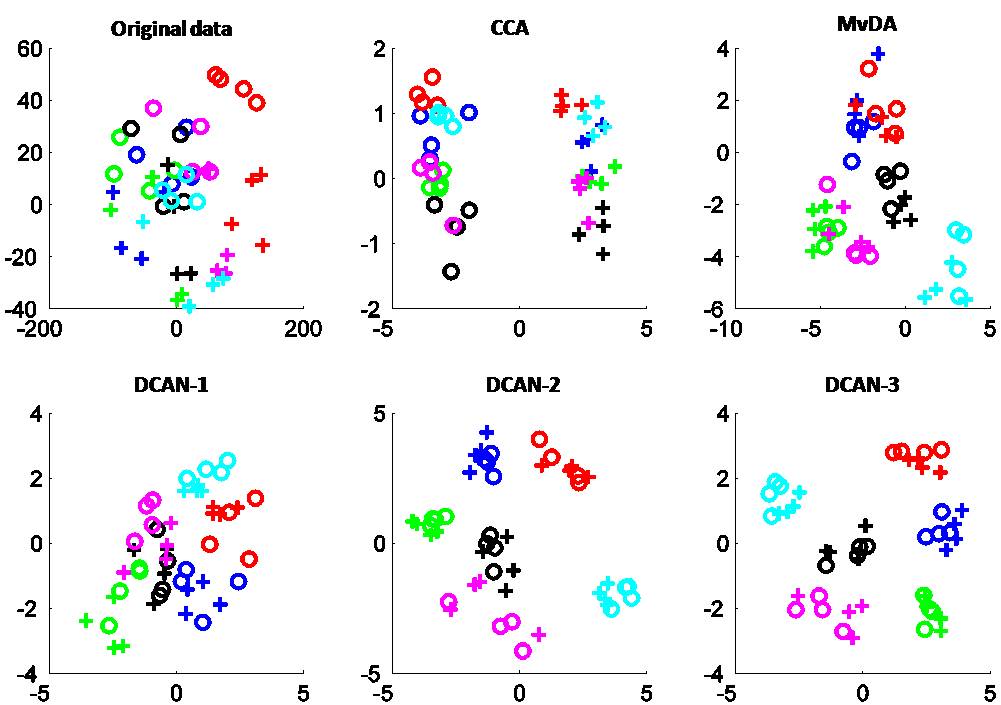}
\end{center}
   \caption{After learning common features by the cross-view methods, we project the features into 2-D space by using the principal two components in PCA. The depicted samples are randomly chosen form Multi-PIE \cite{gross2007cmu} dataset. The ``$\circ$" and ``$+$" points come from two views respectively. Different color points belong to different classes. DCAN-k is our proposed method with a stacked k-layer neural network.}
\label{fig:stack_efficient}
\end{figure}

First, to explicitly illustrate the learnt mapping, we conduct an experiment on Multi-PIE dataset by projecting the learnt common features into a 2-D space with Principal Component Analysis (PCA). As shown in Fig.\ref{fig:stack_efficient}. The classical method CCA can only roughly align the data in the principal directions and the state-of-the-art method MvDA \cite{kan2012multi} attempts to merge two types of data but seems to fail. Thus, we argue that linear transforms are a little stiff to convert data from two views into an ideal common space. The three diagrams below shows that DCAN can gradually separate samples from different classes with the increase of layers, which is just as we described in the above analysis.

Next, we compare our methods with several state-of-the-art methods for the cross-view face recognition task on Multi-PIE data set. Since the images are acquired over seven poses on Multi-PIE data set, in total $7\times 6=42$ comparison experiments need to be conducted. The detailed results are shown in Table \ref{tab:DCAN},where two poses are used as the gallery and probe set to each other and the rank-1 recognition rate is reported. Further, the mean accuracy of all pairwise results for each methods is also reported in Table \ref{tab:Mean_MPIE}.

From Table \ref{tab:Mean_MPIE}, we can find the supervised methods except CSR are significantly superior to CCA due to the use of the label information. And nonlinear methods except Deep CCA are significantly superior to the nonlinear methods due to the use of nonlinear transforms. Compared with FDA, the proposed DCAN with only one layer network can perform better with 1.6\% improvement. With increasing layers, the accuracy of DCAN reaches a climax via stacking three layer networks. The reason of the degradation in DCAN with four layers is mainly the effect of reduced dimensionality, where 10 dimensions are cut out from the above layer network. Obviously, compared with two-view based methods, the proposed DCAN with three layers improves the performance greatly (88.4\% vs. 81.4\%). Besides, MvDA also achieves a considerably good performance by using all samples from all poses. It is unfair to compare these two-view based methods (containing DCAN) with MvDA, because the latter implicitly uses additional five views information except current compared two views. But our method performs better than MvDA, 88.4\% vs. 86.7\%. As observed in Table \ref{tab:DCAN}, three-layer DCAN achieves a largely improvement compared with CCA,FDA,CDFE for all cross-view cases and MvDA for most of cross-view cases. The results are shown in Table \ref{tab:DCAN} and Table \ref{tab:Mean_MPIE}.

\subsection{Photo-Sketch Recognition}

\begin{table}\footnotesize
\setlength{\tabcolsep}{8pt}

\begin{center}
{
    \begin{tabular}{|p{50pt}<{\centering}|p{50pt}<{\centering}|p{50pt}<{\centering}|}
        \hline
        Method & Photo-Sketch & Sketch-Photo\\
        \hline\hline
        CCA\cite{hotelling1936relations} & 0.387 & 0.475\\
        \hline
        KernelCCA\cite{hardoon2004canonical} & 0.466  & 0.570\\
        \hline
        DeepCCA\cite{andrewdeep}& 0.364 & 0.434\\
        \hline
        CDFE\cite{lin2006inter} & 0.456 & 0.476\\
        \hline
        CSR\cite{lei2009coupled} & 0.502 & 0.590\\
        \hline
        PLS\cite{sharma2011bypassing} & 0.486 & 0.510\\
        \hline
        FDA\cite{belhumeur1997eigenfaces} & 0.468 & 0.534\\
        \hline
        MvDA\cite{kan2012multi} & 0.534 & 0.555\\
        \hline\hline
        \textbf{DCAN-1} & 0.535 & 0.555\\
        \hline
        \textbf{DCAN-2} & \textbf{0.603} & 0.613\\
        \hline
        \textbf{DCAN-3} & \textbf{0.601} &\textbf{0.652}\\
        \hline
    \end{tabular}
}
\end{center}

\caption{Evluation on CUFSF database in terms of mean accuracy. DCAN-k means a stacked k-layer network.}

\label{tab:Mean_CUFSF}
\end{table}

Photo-Sketch recognition is conducted on CUFSF dataset. The samples come from only two views, photo and sketch. The comparison results are provided in Table \ref{tab:Mean_CUFSF}. As shown in this table, since only two views can be utilized in this case, MvDA degrades to a comparable performance with those previous two-view based methods. Our proposed DCAN with three layer networks can achieve even better with more than 6\% improvement, which further indicates DCAN benefits from the nonlinear and multi-layer structure.

\textbf{Discussion and analysis:}
The above experiments demonstrate that our methods can work very well even on a small sample size. The reasons lie in three folds:
\begin{enumerate}[(1)]
\item The maximum margin criterion makes the learnt mapping more discriminative, which is a straightforward strategy in the supervised classification task.
\item Auto-encoder approximately preserves the local neighborhood structures. \\
For this, Alain et al. \cite{alain2012regularized} theoretically
prove that the learnt representation by auto-encoder can recover
local properties from the view of manifold. To further validate
that, we employ the first 700 photo images from CUFSF database to
perform the nonlinear self-reconstruction with auto-encoder. With
the hidden presentations, we find the local neighbors with 1,2,3,4,5
neighbors can be preserved with the probability of 99.43\%, 99.00\%,
98.57\%, 98.00\% and 97.42\% respectively. Thus, the use of
auto-encoder intrinsically reduces the complexity of the
discriminant model, which further makes the learnt model better
generality on the testing set.
\item The deep structure generates a gradual model, which makes the learnt transform more robust. With only one layer, the model can't represent the complex data very well. But with layers goes deeper, the coupled networks can learn transforms much more flexible and hence can be allowed to handle more complex data.
\end{enumerate}

\section{Conclusion}\label{sec:Conclusion}

In this paper, we propose a deep learning method, the Deeply Coupled
Auto-encoder Networks(DCAN), which can gradually generate a coupled
discriminant common representation for cross-view object
classification. In each layer we take both local consistency and
discrimination of projected data into consideration. By stacking
multiple such coupled network layers, DCAN can gradually improve the
learnt shared features in the common space. Moreover, experiments in
the cross-view classification tasks demonstrate the superior of our
method over other state-of-the-art methods.


{\small
\bibliographystyle{ieee}
\bibliography{egbib}

\begin{thebibliography}{10}\itemsep=-1pt

\bibitem{akaho2006kernel}
S.~Akaho.
\newblock A kernel method for canonical correlation analysis, 2006.

\bibitem{alain2012regularized}
G.~Alain and Y.~Bengio.
\newblock What regularized auto-encoders learn from the data generating
  distribution.
\newblock {\em arXiv preprint arXiv:1211.4246}, 2012.

\bibitem{andrewdeep}
G.~Andrew, R.~Arora, J.~Bilmes, and K.~Livescu.
\newblock Deep canonical correlation analysis.

\bibitem{belhumeur1997eigenfaces}
P.~N. Belhumeur, J.~P. Hespanha, and D.~J. Kriegman.
\newblock Eigenfaces vs. fisherfaces: Recognition using class specific linear
  projection.
\newblock {\em IEEE Transactions on Pattern Analysis and Machine Intelligence},
  19(7):711--720, 1997.

\bibitem{bengio2013representation}
Y.~Bengio, A.~Courville, and P.~Vincent.
\newblock Representation learning: A review and new perspectives.
\newblock 2013.

\bibitem{bengio2007greedy}
Y.~Bengio, P.~Lamblin, D.~Popovici, and H.~Larochelle.
\newblock Greedy layer-wise training of deep networks, 2007.

\bibitem{chen2012marginalized}
M.~Chen, Z.~Xu, K.~Weinberger, and F.~Sha.
\newblock Marginalized denoising autoencoders for domain adaptation, 2012.

\bibitem{chen2010predictive}
N.~Chen, J.~Zhu, and E.~P. Xing.
\newblock Predictive subspace learning for multi-view data: a large margin
  approach, 2010.

\bibitem{gross2007cmu}
R.~Gross, I.~Matthews, J.~Cohn, T.~Kanade, and S.~Baker.
\newblock The cmu multi-pose, illumination, and expression (multi-pie) face
  database, 2007.

\bibitem{hardoon2004canonical}
D.~R. Hardoon, S.~Szedmak, and J.~Shawe-Taylor.
\newblock Canonical correlation analysis: An overview with application to
  learning methods.
\newblock {\em Neural Computation}, 16(12):2639--2664, 2004.

\bibitem{trevor2001elements}
T.~Hastie, R.~Tibshirani, and J.~J.~H. Friedman.
\newblock The elements of statistical learning, 2001.

\bibitem{hinton2006fast}
G.~E. Hinton, S.~Osindero, and Y.-W. Teh.
\newblock A fast learning algorithm for deep belief nets.
\newblock {\em Neural computation}, 18(7):1527--1554, 2006.

\bibitem{hinton2006reducing}
G.~E. Hinton and R.~R. Salakhutdinov.
\newblock Reducing the dimensionality of data with neural networks.
\newblock {\em Science}, 313(5786):504--507, 2006.

\bibitem{hotelling1936relations}
H.~Hotelling.
\newblock Relations between two sets of variates.
\newblock {\em Biometrika}, 28(3/4):321--377, 1936.

\bibitem{kan2012multi}
M.~Kan, S.~Shan, H.~Zhang, S.~Lao, and X.~Chen.
\newblock Multi-view discriminant analysis.
\newblock pages 808--821, 2012.

\bibitem{kim2007discriminative}
T.-K. Kim, J.~Kittler, and R.~Cipolla.
\newblock Discriminative learning and recognition of image set classes using
  canonical correlations.
\newblock {\em IEEE Transactions on Pattern Analysis and Machine Intelligence},
  29(6):1005--1018, 2007.

\bibitem{le2011optimization}
Q.~V. Le, J.~Ngiam, A.~Coates, A.~Lahiri, B.~Prochnow, and A.~Y. Ng.
\newblock On optimization methods for deep learning, 2011.

\bibitem{lecun1998efficient}
Y.~LeCun, L.~Bottou, G.~B. Orr, and K.-R. M{\"u}ller.
\newblock Efficient backprop.
\newblock In {\em Neural networks: Tricks of the trade}, pages 9--50. Springer,
  1998.

\bibitem{lee2007efficient}
H.~Lee, A.~Battle, R.~Raina, and A.~Y. Ng.
\newblock Efficient sparse coding algorithms, 2007.

\bibitem{lei2009coupled}
Z.~Lei and S.~Z. Li.
\newblock Coupled spectral regression for matching heterogeneous faces, 2009.

\bibitem{li2006efficient}
H.~Li, T.~Jiang, and K.~Zhang.
\newblock Efficient and robust feature extraction by maximum margin criterion.
\newblock {\em Neural Networks, IEEE Transactions on}, 17(1):157--165, 2006.

\bibitem{lin2006inter}
D.~Lin and X.~Tang.
\newblock Inter-modality face recognition.
\newblock pages 13--26, 2006.

\bibitem{ngiam2011multimodal}
J.~Ngiam, A.~Khosla, M.~Kim, J.~Nam, H.~Lee, and A.~Y. Ng.
\newblock Multimodal deep learning, 2011.

\bibitem{nocedal2006numerical}
J.~Nocedal and S.~J. Wright.
\newblock Numerical optimization, 2006.

\bibitem{phillips1998feret}
P.~J. Phillips, H.~Wechsler, J.~Huang, and P.~J. Rauss.
\newblock The feret database and evaluation procedure for face-recognition
  algorithms.
\newblock {\em Image and vision computing}, 16(5):295--306, 1998.

\bibitem{sharma2011bypassing}
A.~Sharma and D.~W. Jacobs.
\newblock Bypassing synthesis: Pls for face recognition with pose,
  low-resolution and sketch, 2011.

\bibitem{sharma2012generalized}
A.~Sharma, A.~Kumar, H.~Daume, and D.~W. Jacobs.
\newblock Generalized multiview analysis: A discriminative latent space, 2012.

\bibitem{vincent2008extracting}
P.~Vincent, H.~Larochelle, Y.~Bengio, and P.-A. Manzagol.
\newblock Extracting and composing robust features with denoising autoencoders,
  2008.

\bibitem{wang2007feature}
F.~Wang and C.~Zhang.
\newblock Feature extraction by maximizing the average neighborhood margin.
\newblock In {\em Computer Vision and Pattern Recognition, 2007. CVPR'07. IEEE
  Conference on}, pages 1--8. IEEE, 2007.

\bibitem{wang2012semi}
S.~Wang, L.~Zhang, Y.~Liang, and Q.~Pan.
\newblock Semi-coupled dictionary learning with applications to image
  super-resolution and photo-sketch synthesis, 2012.

\bibitem{wang2009face}
X.~Wang and X.~Tang.
\newblock Face photo-sketch synthesis and recognition.
\newblock {\em IEEE Transactions on Pattern Analysis and Machine Intelligence},
  31(11):1955--1967, 2009.

\bibitem{xie2012image}
J.~Xie, L.~Xu, and E.~Chen.
\newblock Image denoising and inpainting with deep neural networks, 2012.

\bibitem{yan2007graph}
S.~Yan, D.~Xu, B.~Zhang, H.-J. Zhang, Q.~Yang, and S.~Lin.
\newblock Graph embedding and extensions: a general framework for
  dimensionality reduction.
\newblock {\em IEEE Transactions on Pattern Analysis and Machine Intelligence},
  29(1):40--51, 2007.

\bibitem{zhang2011coupled}
W.~Zhang, X.~Wang, and X.~Tang.
\newblock Coupled information-theoretic encoding for face photo-sketch
  recognition, 2011.

\bibitem{zhao2008maximum}
B.~Zhao, F.~Wang, and C.~Zhang.
\newblock Maximum margin embedding.
\newblock In {\em Data Mining, 2008. ICDM'08. Eighth IEEE International
  Conference on}, pages 1127--1132. IEEE, 2008.

\end{thebibliography}
}

\end{document}